# Hyperbolic Cosine Transformer for LiDAR 3D Object Detection*

Jigang Tong, Fanhang Yang, Sen Yang, Enzeng Dong, Shengzhi Du, Xing Wang, *Member, IEEE,*
and Xianlin Yi

*Abstract*—Recently, Transformer has achieved great success in computer vision. However, it is constrained because the spatial and temporal complexity grows quadratically with the number of large points in 3D object detection applications. Previous point-wise methods are suffering from time consumption and limited receptive fields to capture information among points. In this paper, we propose a two-stage hyperbolic cosine transformer (ChTR3D) for 3D object detection from LiDAR point clouds. The proposed ChTR3D refines proposals by applying cosh-attention in linear computation complexity to encode rich contextual relationships among points. The cosh-attention module reduces the space and time complexity of the attention operation from $O(2N^2d)$ to $O(2Nd^2)$, which is a significant improvement because the number of points ($N$) is usually far greater than the feature dimension ($d$). The traditional softmax operation is replaced by non-negative ReLU activation and hyperbolic-cosine-based operator with re-weighting mechanism. Extensive experiments on the widely used KITTI dataset demonstrate that, compared with vanilla attention, the cosh-attention significantly improves the inference speed with competitive performance. Experiment results show that, among two-stage state-of-the-art methods using point-level features, the proposed ChTR3D is the fastest one.

*Index Terms*—Object detection, segmentation and categorization, deep learning for vision perception.

## I. INTRODUCTION

In recent years, with the development of auto-driving and virtual reality, 3D object detection as an indispensable part of them has received more and more attention from researchers. Point clouds captured by LiDAR sensors are common data modes to represent 3D objects. Unlike RGB images where CNN-like operators are well suited, point clouds are sparse and unordered. To solve this issue, several methods [1][2] convert sparse point clouds into discrete voxels, then apply 3D sparse convolution to extract features from voxels. But these methods will inevitably sacrifice the accurate position information [3]. An important family represented by PointNet [4] directly feeds the raw point clouds into the model through employing permutation invariant operation. However, the point-based methods [4][5] are time-consuming due to repeated calculation, and heavily rely on hand-crafted designs.

Besides of these methods, the Transformer [6] shows great potential for processing point clouds [7][8], because of its inherently permutation invariance. However, the spatial and temporal complexity of attention computation grows quadratically with the size of input point clouds, due to the softmax operation. The main concern is that vanilla transformer attention is intractable for long sequence point clouds.

In this paper, the following contributions are made to the above problems. Firstly, inspired by CosFormer [9], the cosh-attention is proposed to relax the computation bottleneck of point-wise transformer in processing long sequence point clouds. Extensive experiments on the KITTI dataset demonstrate that, in 3D object detection tasks, compared with vanilla attention, the cosh-attention significantly improves the inference speed with similar competitive performance. Secondly, the Hyperbolic Cosine Transformer (ChTR3D) is proposed, which is an end-to-end two-stage framework for LiDAR 3D object detection based on cosh-attention. Inspired by DETR [10] and CT3D [3], the proposed ChTR3D generates proposals at the first stage by applying 3D sparse convolution to extract features from voxels. Then at the second stage, the cosh-attention-based encoder-decoder architecture is used in linear calculation complexity to extract point-wise features from raw points for refinement. Comparing with other point-wise and voxel convolutions suffering from limitations including parameter optimization and receptive field, the proposed point-wise transformer-based architecture is more flexible and has a larger receptive field. Our work provides a feasible scheme for the study of direct applying attention operation on point clouds.

The rest of the paper is organized as follows. Section 2 introduces the related work. Section 3 discusses the limitations of existing point-wise transformers in point cloud tasks. Then according to these limitations, the ChTR3D is proposed. The fourth section shows the performance comparison and makes a brief analysis between the proposed ChTR3D and other state-of-the-art approaches on the KITTI dataset.

## II. RELATED WORK

### A. Point-based LiDAR 3D Object Detection

The research on point-based 3D object detection has ushered in rapid development, since the PointNet [4] and PointNet++ [5] directly using the raw point clouds as input. F-PointNet [11] is the first 3D object detection method using PointNet [4]. It generated proposals by 2D object detection based on RGB images. Then, PointRCNN [12] used PointNet++ [5] as the backbone to segment the whole point clouds, and generated proposals based on the separated foreground points. The Sparse-to-Dense (STD) method [13] seeded a spherical anchor for each point to generate proposals, and presented PointsPool layer for feature extraction. The 3D-SSD [14] proposed a single-stage detector based on a novel point cloud sampling algorithm using feature distance, which replaces the time-consuming feature propagation.

## B. Grid-based LiDAR 3D Object Detection

The grid-based methods transform the unordered sparse point clouds to regular either 3D voxels [1][2][15][16] or 2D bird-view maps [17]-[19], from which features are extracted by CNN-like operators. The SECOND [1] proposed an efficient one-stage 3D object detection method with high recall rate by 3D sparse convolutions. Voxel R-CNN [2] presented voxel region of interest (RoI) pooling that extracts RoI features directly from voxels for further refinement.

## C. Point-Voxel-based LiDAR 3D Object Detection

The SA-SSD [20] presented an auxiliary network based point-level supervision training. The two-stage PV-RCNN [21] firstly sampled keypoints through farthest point sampling (FPS), and then refined the proposals at the second stage, by combining both multi-scale voxel features and keypoint features extracted using PointNet [4].

## D. Transformer for Point Clouds

Recently, the Transformer [6] achieved great success in image classification [22] and 2D object detection [10], where the attention got intensive emphasis. For point cloud classification and segmentation [7][8][23][24], the point transformer [7] was equipped with a self-attention layer for point clouds. The Point Cloud Transformer (PCT) [8] used an offset-attention and normalization mechanism to boost the effectiveness. For 3D object detection tasks, the CT3D [3] presented a two-stage object detector, where the vanilla self-attention was used to encode the channel-wise features in RoI for further refinement. The Voxel Transformer (VoTr) [25] proposed a transformer-based backbone with high recall rate to replace sparse convolutions.

## III. METHOD

In this section, we will propose the Hyperbolic Cosine Transformer (ChTR3D), a new two-stage 3D object detection framework. Generally, 3D voxel sparse CNN can be used to efficiently generate high-quality proposals. But the voxel-based proposal refinement methods [1][2] suffer the difficulties of extra hyper-parameter optimization [3] caused by the voxel sensitivity. Therefore, we firstly use voxel-based region proposal network (RPN) to generate high-quality proposals effectively, and then apply point-wise features with retaining fine-grained localization to refine proposals. As shown in Fig. 1, ChTR3D includes 3D backbone network and RPN for generating proposal at the first stage, then the hyperbolic-cosine-based transformer for refinement and detect head for predicting final bounding boxes at the second stage.

## A. 3D Proposal Generation

Due to the efficiency and high-recall performance, the 3D CNN SECOND [1] is adopted as default RPN of ChTR3D to generate the proposals. Given the point clouds $P$ with $3$-dimension spatial coordinates and $C$-dimension point features, the proposals generated by RPN consists of the following information: center coordinate $p^{prop} = [x^{prop}, y^{prop}, z^{prop}]$, length $l^{prop}$, width $w^{prop}$, height $h^{prop}$ and orientation $\theta^{prop}$.

## B. Preliminary

In this subsection, more details and limitations of vanilla transformer attention in point cloud tasks will be discussed. And then, according to these limitations, we provide solutions to pave the way for the cosh-attention proposed later.

*Vanilla Softmax Attention:* Given a 3D point cloud scene $x$ with $N$ points and $d$-dimension feature, i.e. $x \in \mathbb{R}^{N \times d}$. A transformer block $\mathcal{T}: \mathbb{R}^{N \times d} \to \mathbb{R}^{N \times d}$ with input $x$ can be summarized as (1).

$$\mathcal{T}(x) = \mathcal{F}(\mathcal{A}(x) + x), \qquad (1)$$

where $\mathcal{F}$ denotes a feedforward network (FFN) with residual connection, $\mathcal{A}$ is the self-attention operation to calculate the attention matrix A for input $x$. $\mathcal{A}$ has quadratic computation complexity with respect to $N$. Thus, $\mathcal{A}$ becomes the calculation bottleneck of $\mathcal{T}$ for long-range point clouds. Specifically, in $\mathcal{A}$, we have query ($Q \in xW_q$), key ($K \in xW_k$), and value ($V \in xW_v$), where the $W_q, W_k, W_v \in \mathbb{R}^{d \times d}$ are linear projection matrices. The output $\mathcal{O} \in \mathbb{R}^{N \times d}$ of $\mathcal{A}(x)$ can be described in (2).

$$\mathcal{O} = \mathcal{A}(x) = [\mathcal{O}_1, \mathcal{O}_2, \dots, \mathcal{O}_N]^T, \mathcal{O}_i = \sum_j \frac{\mathcal{S}(Q_i, K_j)}{\sum_j(Q_i, K_j)} V_j, \qquad (2)$$

where $\mathcal{O}_i$ represents the $i$-th row of output attention matrix $\mathcal{O}$, and $\mathcal{S}(\cdot)$ is the computation similarity function. If $\mathcal{S}(Q, K) = \exp(QK^T)$ and the scaling $\sqrt{d}$ is ignored, then (2) becomes the vanilla transformer attention using softmax operator as shown in (3).

$$\mathcal{A}(Q, K, V) = [Softmax(QK^T) \times V]. \qquad (3)$$

As illustrated in the left part of Fig. 2, $N$ is the size of point clouds, $d$ is the feature dimension. In most cases, $N \gg d$. The computation complexity of $\mathcal{O}_i$ is $O(Nd)$. The total computation complexity of $\mathcal{O}$ is $O(2N^2d)$, which grows quadratically with respect to the size of input point clouds. While the linearized self-attention operation achieves a linear calculation complexity $O(N)$.

*Linearized Self-attention:* Inspired by CosFormer [9], the key to reduce the quadratic computation complexity of vanilla transformer attention is replacing the non-linear and indecomposable softmax operation by a linear operation with decomposable non-linear re-weighting mechanism. Thus, the similarity $\mathcal{S}(\cdot)$ in (2) can be decomposed as:

$$\mathcal{S}(Q_i, K_j) = \emptyset(Q_i) \emptyset(K_j)^T. \qquad (4)$$

Then, the linear complexity of attention operation can be achieved through the matrix product property:

$$[\emptyset(Q_i)\emptyset(K_j)^T]V = \emptyset(Q_i)[\emptyset(K_j)^T V]. \qquad (5)$$

As shown in the right part of Fig. 2, if $\emptyset(K_j)^T V \in \mathbb{R}^{d \times d}$ is calculated first, then multiplied by $Q \in \mathbb{R}^{N \times d}$ on the left side, it will only bring the calculation complexity of $O(2Nd^2)$, rather than the calculation complexity with $O(2N^2d)$ of vanilla transformer using softmax operation.

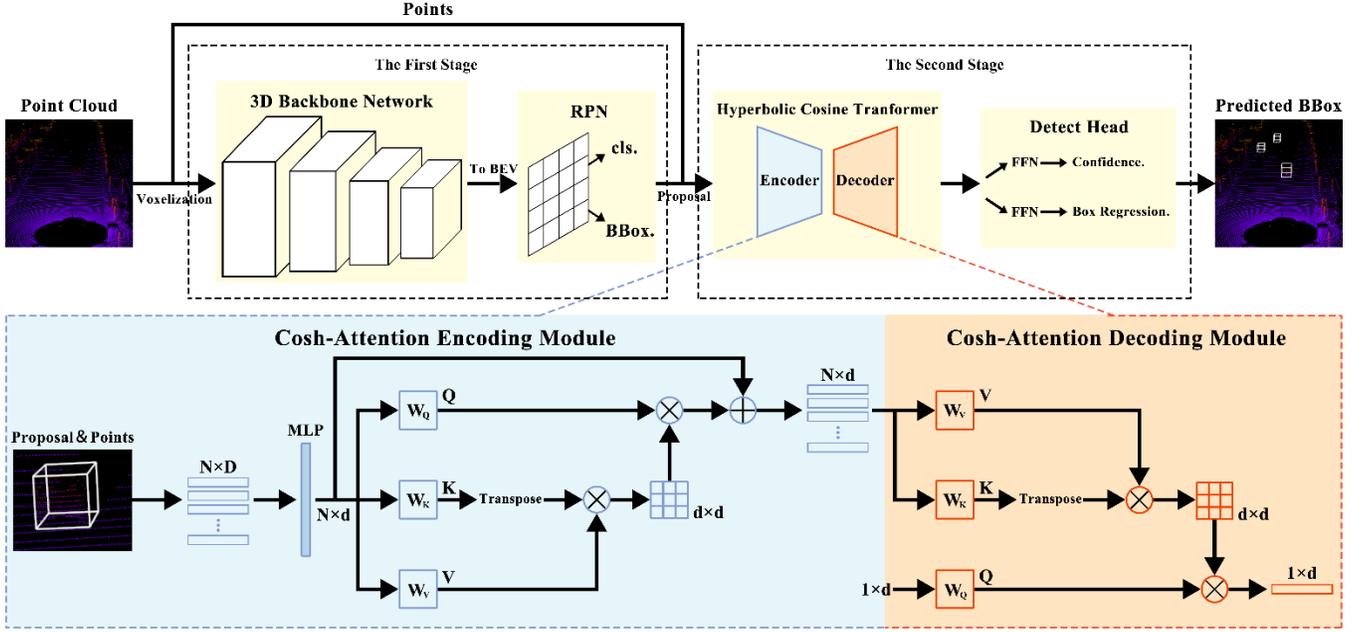

Figure 1. The overview architecture of ChTR3D, where the BEV means Birds-Eye-View. In ChTR3D, the point clouds is firstly voxelized and fed into the 3D backbone network with sparse convolution and RPN to generate proposals. Then the cosh-attention encoding-decoding module is used to convert point-wise features into a final global proposal representation for further confidence prediction and bounding box regression.

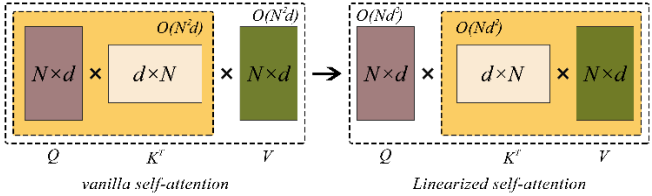

Figure 2. Illustration of computation complexity for vanilla self-attention(left) and linearized self-attention(right).

Generally, the size of point clouds $N$ is much larger than the feature dimension $d$ in 3D object detection tasks. Therefore, it is very important for point-wise transformer to achieve a linear computation complexity when it is applied to point cloud tasks.

### C. The Proposed Cosh-attention

According to CosFormer [9] and other papers [26]-[28], the softmax operator has two important properties that play core roles in its performance:

- It provides a non-linear re-weighting mechanism to concentrate the distribution of attention weights and therefore stabilize the training process.
- It ensures all values in the attention matrix A to be non-negative.

For the non-negative property, CosFormer [9] has proved that ReLU activation can well replace softmax operation to effectively eliminate negative values in the attention matrix A.

The proposed cosh-attention with non-linear re-weighting mechanism conforms to the operation of linearized attention and stabilizes the attention weights, where the following two critical components are included.

*Non-negative Linear Projection:* To ensure that the values of attention matrix A elements are all non-negative and avoid re-weighting mechanism aggregating negative-correlated information, we also adopt the ReLU activation to eliminate the negative values of $Q, K$ and $V$, as shown in (6).

$$\mathcal{S}(Q,K) = s[ReLU(Q), ReLU(K)] = s(Q', K'), \quad (6)$$

where $s(\cdot)$ is a decomposable cosh-based function in our method, which computes the similarity between $Q'$ and $K'$. As done in CosFormer [9], dot-product-like operation, e.g. $s(x,y) = xy^T$, $x, y \in \mathbb{R}^{1 \times d}$ followed by a row-wise normalization is adopted to calculate attention matrices.

*Cosh-based Re-weighting Mechanism:* For the re-weighting property of softmax operation, we use hyperbolic-cosine-based re-weighting scheme as shown in (7) to concentrate the distribution of attention weights for stabilizing training.

$$f(x) = 2 - \cosh(x). \quad (7)$$

The hyperbolic-cosine weights can be decomposed into three summations, which satisfies the requirements of the operation of linearized attention. Specifically, by combining with (6) and (7), the similarity function operation with cosh-based re-weighting can be defined in (8).

$$s(Q'_i, K'_j) = Q'_i {K'_j}^T \left[2 - \cosh\left(a \times \frac{i-j}{M}\right)\right]. \quad (8)$$

Equation (8) can be further decomposed as shown in (9).

$$Q'_i {K'_j}^T \left[2 - \cosh\left(a \times \frac{i-j}{M}\right)\right]$$
$$= Q'_i {K'_j}^T \left[2 - \cosh\left(\frac{ai}{M}\right)\cosh\left(\frac{aj}{M}\right) + \sinh\left(\frac{ai}{M}\right)\sinh\left(\frac{aj}{M}\right)\right]$$

$$= 2Q'_i K'^T_j - \left[Q'_i \cosh\left(\frac{ai}{M}\right)\right]\left[K'^T_j \cosh\left(\frac{aj}{M}\right)\right]$$
$$+ \left[Q'_i \sinh\left(\frac{ai}{M}\right)\right]\left[K'^T_j \sinh\left(\frac{aj}{M}\right)\right], \quad (9)$$

where $a$ is a hyperparameter, and $i,j = 1,\ldots,N, M \geq N$, $-1 < \frac{i-j}{M} < 1$. If $a \leq 1.3169$, then $2 - \cosh\left(a \times \frac{i-j}{M}\right)$ is non-negative. The cosh-based re-weighting mechanism introduced in (7) provides different weights in the manner of more favorite to the ones closer to the current one in the sequence (such as for the case that $i$ and $j$ are close to each other in (8)), which may not be suitable for unordered point clouds. As shown in Fig. 3, a series of ablation experiments is designed to make the distribution of attention weights more concentrated by changing the value of $a$. And as shown in Table 3, we found that the weight distribution does not significantly affect the final performance on the KITTI val set. Then, considering (6), let

$$Q_i^{\cosh} = Q'_i \cosh\left(\frac{ai}{M}\right), \quad (10)$$
$$Q_i^{\sinh} = Q'_i \sinh\left(\frac{ai}{M}\right), \quad (11)$$
$$K_j^{\cosh} = K'_j \cosh\left(\frac{aj}{M}\right), \quad (12)$$
$$K_j^{\sinh} = K'_j \sinh\left(\frac{aj}{M}\right). \quad (13)$$

The output of the proposed cosh-attention at the $i^{th}$ point cloud sequence position is:

$$\mathcal{O}_i = \frac{\sum_{j=1}^N f(Q'_i, K'_j) V_j}{\sum_{j=1}^N f(Q'_i, K'_j)}$$
$$= \frac{2\sum_{j=1}^N Q'_i [(K'_j)^T V_j] - \sum_{j=1}^N Q_i^{\cosh}[(K_j^{\cosh})^T V_j] + \sum_{j=1}^N Q_i^{\sinh}[(K_j^{\sinh})^T V_j]}{2\sum_{j=1}^N Q'_i (K'_j)^T - \sum_{j=1}^N Q_i^{\cosh}(K_j^{\cosh})^T + \sum_{j=1}^N Q_i^{\sinh}(K_j^{\sinh})^T}. \quad (14)$$

From (14), the cosh-attention operation achieves the output in (15) with a linear complexity.

$$\mathcal{O} = \mathcal{S}(Q,K)V = (2QK - Q^{\cosh}K^{\cosh} + Q^{\sinh}K^{\sinh})V$$
$$= 2Q(KV) - Q^{\cosh}(K^{\cosh}V) + Q^{\sinh}(K^{\sinh}V). \quad (15)$$

*D. Hyperbolic Cosine Transformer for 3D Object Detection*

In order to refine the proposal with fine-grained point-wise features, inspired by DETR [10], we design an encoder-decoder architecture based on cosh-attention. Specifically, in the encoding module, we adopt proposal-to-point strategy [3] to encode the proposal information into point-wise features. Then, the proposed cosh-self-attention is used to encode the rich contextual information among the proposal-aware points with linear computation complexity for further refinement. In the decoder module, all point-wise features are converted into a final global proposal representation by cosh-cross-attention. Finally, the proposal representation vector is used to predict the final bounding boxes by two separate FFNs.

*Cosh-based-attention Encoder Module:* Following the principle of refining proposals to fit the corresponding ground truth bounding boxes by covering object points as accurate as possible, we delineate a scaled RoI area in point clouds based on proposals generated at the first stage. Specifically, the spherical RoI area is defined by the distance from the proposal center point to the corner point as the radius, $r_{RoI} = \alpha \sqrt{\left(\frac{l^c}{2}\right)^2 + \left(\frac{w^c}{2}\right)^2 + \left(\frac{h^c}{2}\right)^2}$, where the scaling factor $\alpha$ is a hyperparameter. Then $N$ points are randomly selected from RoI area for subsequent processing, which are denoted by $\mathcal{N} = \{p_1, \ldots, p_N\}$, where $p_i$ is the coordinate of the $i$-th sampled raw point.

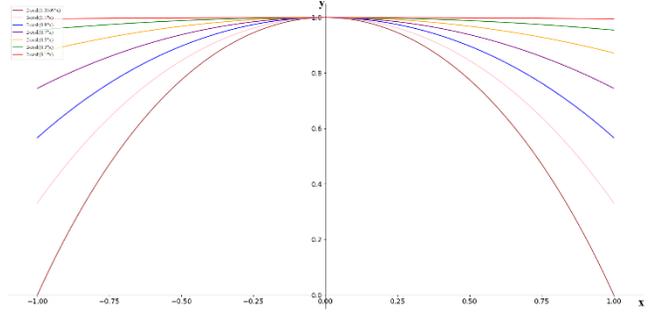

Figure 3. Controlling the weights distribution in the point cloud sequence by changing the value of $a$.

We adopt the proposal-to-point strategy [3] to calculate the 3D relative coordinates between sampled points and corresponding proposal points as point-wise features. The proposal points include a center and eight corner points. Specifically, the relative coordinates are calculated in (16).

$$\Delta p_i^j = p_i - p^j, j = 1,2,\ldots,9, \quad (16)$$

where $p_i$ is the coordinate of sampled raw point, $p^j$ is the reference point which can be chosen from the coordinates of the corresponding proposal center and eight corner points. As shown in Fig. 4, the geometry information of the proposal $(x^{\text{prop}}, y^{\text{prop}}, z^{\text{prop}}, l^{\text{prop}}, w^{\text{prop}}, h^{\text{prop}}, \theta^{\text{prop}})$ is encoded to the localization features of each point by the above calculation. The proposal-aware features of each sampled point are shown in (17).

$$\mathcal{P}_i = [\Delta p_i^1, \Delta p_i^2, \ldots, \Delta p_i^9, p_i^C] \in \mathbb{R}^{1 \times (27+r)}. \quad (17)$$

For KITTI dataset, the point cloud data are recorded in the form of 3D coordinates $(x, y, z)$ and reflectance information $r$. Therefore $p_i^C \in \mathbb{R}^{1 \times 1}$ and $\mathcal{P}_i \in \mathbb{R}^{1 \times 28}$. And then, $\mathcal{P}_i$ is mapped into a high-dimension embedding by multi-layer perceptron (MLP) with two linear layers and one ReLU activation in (18).

$$f_i = \mathcal{W}(\mathcal{P}_i) \in \mathbb{R}^{1 \times d}, \quad (18)$$

where $\mathcal{W}(\cdot)$ denotes MLP, and $d = 64$ in this work.

In this way, the embedded point features are processed by multi-head cosh-self-attention and subsequent FFN to encode spatial contextual dependencies in proposal for refining point-wise features. Specifically, denoting the $d$-dimension point-wise features as

$$\mathcal{X} = [f_1^T, f_2^T, \ldots, f_N^T]^T \in \mathbb{R}^{N \times d}. \quad (19)$$

We have

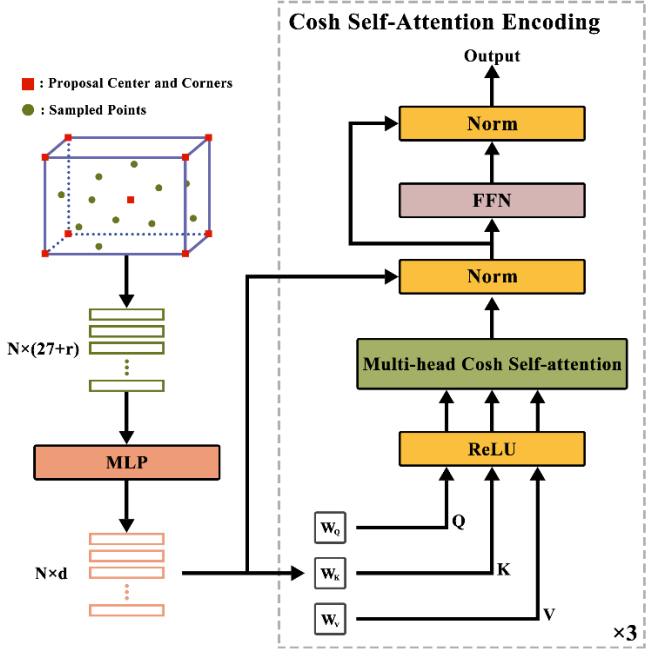

Figure 4. Cosh-self-attention encoding module. The relative coordinates of all the sampled raw points and corresponding proposal points are used as the position features. Then, the multi-head cosh-self-attention is used to encode the sampled proposal-aware point-wise features.

$$\begin{cases} Q \in \mathcal{X}W_q \\ K \in \mathcal{X}W_k \\ V \in \mathcal{X}W_v \end{cases}, \quad (20)$$

where $W_q, W_k, W_v \in \mathbb{R}^{d \times d}$ are linear projection matrices. Then, combining with (6), the $Q, K, V$ are processed by the ReLU activation with non-negative property as follows.

$$\begin{cases} Q' = ReLU(Q) \\ K' = ReLU(K) \\ V' = ReLU(V) \end{cases}. \quad (21)$$

In a H-head attention situation, the $Q', K'$ and $V'$ are divided as shown in (22).

$$\begin{cases} Q' = [Q'_1, \dots, Q'_H] \\ K' = [K'_1, \dots, K'_H] \\ V' = [V'_1, \dots, V'_H] \end{cases}, \quad (22)$$

where $Q'_h, K'_h, V'_h \in \mathbb{R}^{N \times d'}, h = 1, \dots, H, d' = \frac{d}{H}$. Using $Q'_{hi}, K'_{hi}, V'_{hi}$ to denote the $i$-th row of $Q'_h, K'_h, V'_h$ (of the $i$-th point in a point cloud sequence), respectively, and $i, j = 1, \dots, N$. Let

$$\begin{cases} Q^{\cosh}_{hi} = Q'_{hi} \cosh\left(\frac{ai}{M}\right) \\ Q^{\sinh}_{hi} = Q'_{hi} \sinh\left(\frac{ai}{M}\right) \\ K^{\cosh}_{hj} = K'_{hj} \cosh\left(\frac{aj}{M}\right) \\ K^{\sinh}_{hj} = K'_{hj} \sinh\left(\frac{aj}{M}\right) \end{cases}. \quad (23)$$

Then the output of the $h$-head cosh-self-attention for the $i$-th point is

$$\mathcal{O}_{hi} = \frac{2Q'_{hi} \sum_{j=1}^{N} \left[(K'_{hj})^T V_{hj}\right] - Q^{\cosh}_{hi} \sum_{j=1}^{N} \left[(K^{\cosh}_{hj})^T V_{hj}\right] + Q^{\sinh}_{hi} \sum_{j=1}^{N} \left[(K^{\sinh}_{hj})^T V_{hj}\right]}{2Q'_{hi} \sum_{j=1}^{N} (K'_{hj})^T - Q^{\cosh}_{hi} \sum_{j=1}^{N} (K^{\cosh}_{hj})^T + Q^{\sinh}_{hi} \sum_{j=1}^{N} (K^{\sinh}_{hj})^T}. \quad (24)$$

Therefore, the proposed ChTR3D encodes the information among points with linear calculation complexity, the output of $h$-head cosh-self-attention is shown in (25).

$$\begin{aligned} \mathcal{O}_h &= \mathcal{A}(Q_h, K_h, V_h) \\ &= 2Q_h(K_h^T V_h) - Q_h^{\cosh}\left(K_h^{\cosh^T} V_h\right) \\ &\quad + Q_h^{\sinh}\left(K_h^{\sinh^T} V_h\right), h = 1, \dots, H. \end{aligned} \quad (25)$$

The total output of multi-head cosh-self-attention can be presented in (26).

$$\mathcal{M}(Q, K, V) = Concat(\mathcal{O}_h)W^O, h = 1, \dots, H, \quad (26)$$

where $W^O \in \mathbb{R}^{d \times d}$ is linear projection layer. Applying FFN and residual connection, the result can be shown in (27).

$$\mathcal{T}(\mathcal{X}) = \mathcal{N}(\mathcal{F}\{\mathcal{N}[\mathcal{M}(Q, K, V)]\}), \quad (27)$$

where $\mathcal{N}(\cdot)$ denotes the residual connection and normalization layer, $\mathcal{F}(\cdot)$ is an FFN with two linear projection layers and one ReLU layer. It is noteworthy that position embedding is not explicitly used in the proposed ChTR3D, because the point-wise features already contain the position information. Like CT3D [3], a stack of 3 identical cosh-self-attention encoding modules is used in ChTR3D.

*Cosh-cross-attention Decoder Module:* Unlike [10] where the M query embeddings were used, the proposed the multi-head cosh-cross-attention module only contains a single query embedding. The zero initialized query embedding and the key-value embeddings from the encoder are used to decode all the point-wise features into a final global proposal representation by multi-head cosh-cross-attention for further processing.

*E. Detection Head and Training Targets*

The representation vector from previous steps is fed into two separate FFNs, which predict the confidence and box residual, respectively. In the confidence prediction branch, setting the Intersection over Union (IoU) between proposal and ground-truth box as training targets. According to the IoU, as done in [12][21][29], the confidence prediction target is assigned as shown in (28).

$$c^t = min\left[1, max\left(0, \frac{IoU^{prop} - \alpha_B}{\alpha_F - \alpha_B}\right)\right], \quad (28)$$

where $\alpha_F$ and $\alpha_B$ denote the thresholds of foreground and background, respectively. $IoU^{prop}$ is the IoU calculated by 3D proposal and corresponding ground-truth box. In the box regression branch, we follow CT3D [3] to present regression targets (superscript t) by the proposals (superscript prop) and corresponding ground-truth boxes (superscript g), given by:

$$\begin{cases} x^t = \dfrac{x^g - x^{\text{prop}}}{d_{\text{diag}}} \\ y^t = \dfrac{y^g - y^{\text{prop}}}{d_{\text{diag}}} \\ z^t = \dfrac{z^g - z^{\text{prop}}}{h^{\text{prop}}} \\ l^t = \log\left(\dfrac{l^g}{l^{\text{prop}}}\right) \\ w^t = \log\left(\dfrac{w^g}{w^{\text{prop}}}\right) \\ h^t = \log\left(\dfrac{h^g}{h^{\text{prop}}}\right) \\ \theta^t = \theta^g - \theta^{\text{prop}} \end{cases} \quad (29)$$

where $d_{\text{diag}}$ denotes the diagonal length of the bottom of the 3D proposal.

*F. Loss Function*

The end-to-end training strategy is used in ChTR3D, which includes region proposal network loss $\mathcal{L}_{\text{RPN}}$ and proposal refinement loss $\mathcal{L}_{\text{rcnn}}$. The $\mathcal{L}_{\text{total}}$ is calculated by

$$\mathcal{L}_{\text{total}} = \mathcal{L}_{\text{RPN}} + \mathcal{L}_{\text{rcnn}}, \quad (30)$$

where $\mathcal{L}_{\text{rcnn}}$ is composed of IoU-guided confidence prediction loss $\mathcal{L}_{\text{iou}}$ and bounding box regression $\mathcal{L}_{\text{reg}}$.

$$\mathcal{L}_{\text{rcnn}} = \mathcal{L}_{\text{iou}} + \mathcal{L}_{\text{reg}}, \quad (31)$$

where $\mathcal{L}_{\text{iou}}$ is calculated for predicted confidence by binary cross-entropy loss [18][29].

$$\mathcal{L}_{\text{iou}} = -c^t \log(c) - (1 - c^t)\log(1 - c), \quad (32)$$

where $c$ is the predicted confidence by the FFN, $c^t$ is the confidence training target calculated by the 3D proposal and corresponding ground-truth box. Moreover, the bounding box regression loss $\mathcal{L}_{\text{reg}}$ [1][18] is

$$\mathcal{L}_{\text{reg}} = \mathbb{I}(\text{IoU} \geq \alpha_R) \sum_{r \in x,y,z,l,w,h,\theta} \mathcal{L}_{\text{smooth-L1}}(r, r^t), \quad (33)$$

where $\mathbb{I}(\text{IoU} \geq \alpha_R)$ indicates that only proposals with $\text{IoU} \geq \alpha_R$ contribute to the regression loss.

## IV. EXPERIMENTS

*A. Dataset*

The widely used KITTI dataset [30] includes 7481 training samples and 7518 test samples, where the labeled data are divided into the training set with 3712 samples and validation set with 3769 samples for experiments [31].

*B. Implementation of Experiments*

*RPN:* The SECOND [1] is used as the backbone network to generate high-quality proposals at the first stage. For the KITTI dataset, as the same setting of CT3D [3] and PV-RCNN [21], the detection ranges of X, Y, Z axis are set as $(0, 70.4), (-40, 40), (-3, 1)$, and the voxel size is $(0.05m, 0.05m, 0.1m)$ in (X-axis, Y-axis, Z-axis), respectively. $\mathcal{L}_{\text{RPN}}$ for proposal generation at the first stage includes the Focal-Loss-guided classification branch and the Smooth-L1-Loss-guided regression branch. For more details, please refer to the OpenPCDet toolbox framework [32].

*Training Details:* Our ChTR3D framework is trained end-to-end 100 epochs on a single NVIDIA 1080Ti GPU for KITTI Dataset with the Adam optimizer. The initial learning rate and batch size are set to 0.001 and 6, respectively. Besides, we adopt cosine annealing and one-cycle policy to update learning rate. For the cosh-self-attention encoder module, we set scaling factor $\alpha = 1.1$, and 256 points (i.e. $N = 256$) are randomly selected from RoI area for subsequent encoding. For training targets, we set the foreground threshold $\alpha_F = 0.75$, the background threshold $\alpha_B = 0.25$, and the regression threshold $\alpha_R = 0.55$. During training, we randomly select 128 proposals to calculate the IoU-guided confidence loss, while 64 $(IoU \geq \alpha_R)$ proposals are selected to calculate the regression loss. At the inference stage, we choose 100 proposals with the highest confidence for final prediction.

*C. KITTI Dataset Results*

Table 1 shows the performance comparisons between our ChTR3D and other state-of-the-art approaches for the common used "car" category on the official KITTI test server. The test results (i.e. average precision (AP)) are calculated with 0.7 IoU threshold and 40 recall positions on easy, moderate and hard levels by KITTI official server. According to these results, the proposed ChTR3D achieves the best performance on easy level and approaches the best on moderate and hard levels among methods using point-level features (including point-based and point-voxel-based methods). Compared with CT3D [3] and PV-RCNN [21], which also use SECOND [1] as the RPN, ChTR3D achieves the best performance for the "car" category.

TABLE I. PERFORMANCE COMPARISONS WITH THE OTHER PREVIUS APPROACHES ON THE OFFICIAL KITTI TEST SERVER

| Type | Method | Easy | Mod. | Hard | Time(ms) |
|---|---|---|---|---|---|
| Single-stage | *Voxel-based:* | | | | |
| | VoxelNet[15] | 77.82 | 64.17 | 57.51 | - |
| | SECOND[1] | 84.65 | 75.96 | 68.71 | 33 |
| | PointPillars[19] | 82.58 | 74.31 | 68.99 | 16 |
| | SA-SSD[20] | 88.75 | 79.79 | 74.16 | 40 |
| | CIA-SSD[33] | 89.59 | 80.28 | 72.87 | 30 |
| | *Point-based:* | | | | |
| | 3DSSD[15] | 88.36 | 79.57 | 74.55 | 40 |
| Two-stage | *RGB+LiDAR:* | | | | |
| | F-PointNet[11] | 82.19 | 69.79 | 60.59 | 17 |
| | CLOCS[34] | 89.16 | **82.28** | 77.23 | 100 |
| | SRIF-RCNN[35] | 88.45 | 82.04 | 77.54 | 95 |
| | *Voxel-based:* | | | | |
| | Voxel R-CNN[2] | **90.90** | 81.62 | 77.06 | 40(38*) |
| | VoTr-TSD[25] | 89.90 | 82.09 | **79.14** | 70(200*) |
| | VoxSeT[36] | 88.53 | 82.06 | 77.46 | 33 |
| | *Point-based:* | | | | |
| | PointRCNN[12] | 86.96 | 75.64 | 70.70 | 100(125*) |
| | STD[13] | 87.95 | 79.71 | 75.09 | 80 |
| | *PV-based:* | | | | |
| | PV-RCNN[21] | 90.25 | 81.43 | 76.82 | 80(99*) |
| | Pyramid-PV[37] | 88.39 | 82.08 | 77.49 | 70(155*) |
| | E^2-PV-RCNN[38] | 88.33 | 81.70 | 77.20 | 100 |
| | BADet[39] | 89.28 | 81.61 | 76.58 | 140 |
| | CT3D[3] | 87.83 | 81.77 | 77.16 | 70(84*) |
| | ChTR3D(ours) | 90.43 | 82.02 | 77.42 | 61* |

* denotes the inference speed on our server.

TABLE II. ABLATION STUDIES FOR COSH-ATTENTION AND VANILLA ATTENTION ON THE KITTI VALIDATION SET WITH AP CALCULATED BY 11 AND 40 RECALL POSITIONS

| Method | Operation Space | Feature Dimension | IoU = 0.7 (R40) | | | IoU = 0.7 (R11) | | | Runtime(ms) |
|---|---|---|---|---|---|---|---|---|---|
| | | | Easy | Mod. | Hard | Easy | Mod. | Hard | |
| CT3D[3] | Channel-wise | $N=256, d=256$ | **92.66** | 85.08 | **83.47** | **89.84** | 84.23 | **79.61** | 84 |
| CT3D+V.A. | Point-wise | $N=256, d=256$ | 92.18 | 84.98 | 82.81 | 89.08 | 84.89 | 78.68 | 85 |
| CT3D+V.A. | Point-wise | $N=256, d=64$ | 91.57 | 84.99 | 82.86 | 88.54 | 84.91 | 78.70 | 70 |
| CT3D+Cosh.A. | Point-wise | $N=256, d=64$ | 91.93 | **85.09** | 83.09 | 88.86 | **85.08** | 78.83 | 63 |

"V.A." and "Cosh.A." stand for the vanilla attention and Cosh-attention used in the ablation studies, respectively.

Besides, as shown in Fig. 5, we visualize the detection results of CT3D [3] and ChTR3D on the KITTI test set. Compared with CT3D, ChTR3D presents more reasonable detection results. The improvement is relevant to the sampled points being mostly interior points of foreground instances from a smaller spherical RoI in ChTR3D, rather than a larger cylindrical RoI in CT3D.

Moreover, in order to fairly compare the inference speed, we run other methods on the KITTI validation set under the same calculation conditions. As shown in Table 1 (superscript *), compared with other two-stage methods using point-level features (i.e. point-based methods and PV-based methods) to refine proposal, the ChTR3D not only achieves the competitive performance, but also exceeds the inference speed of other methods because of the linear computation complexity as discussed in Section III.B.

*D. Ablation Studies*

In this subsection, we conduct a series of ablation experiments to verify the effectiveness of the proposed cosh-attention. For the reliability, we take the AP average of the last 10 training epochs with 0.7 IoU threshold, 11 and 40 recall positions for the "car" category.

*Cosh-attention versus Vanilla Attention:* We use the proposed cosh-attention and vanilla attention to replace the channel-wise transformer attention in CT3D [3], which is also a two-stage transformer-based framework using point-level features to refine proposals. As shown in Table 2, for the consistent size of input points (i.e. $d=64$), compared with vanilla attention in point-wise method, cosh-attention significantly improves the inference speed and achieves better performance. Compared with the regular point-wise transformer on different sizes of input points, the performance on easy level of our ChTR3D is slightly worse. The reason is that the easy-level objects usually have much larger number of points. Therefore, in these cases, the regular point-wise transformer ($i.e. d=256$) with more parameters performs better than cosh-attention.

*Cosh-based Re-weighting versus Unordered Point Clouds:* For the proposed cosh-based re-weighting mechanism (7) respect to providing different weights in the sequence, which may not be suitable for the unordered point clouds. Thus, we design ablation experiments to make the distribution of attention connections more concentrated by changing the value of $a$. As shown in Table 3, the locality bias caused by different values of $a$ does not significantly affect the performance.

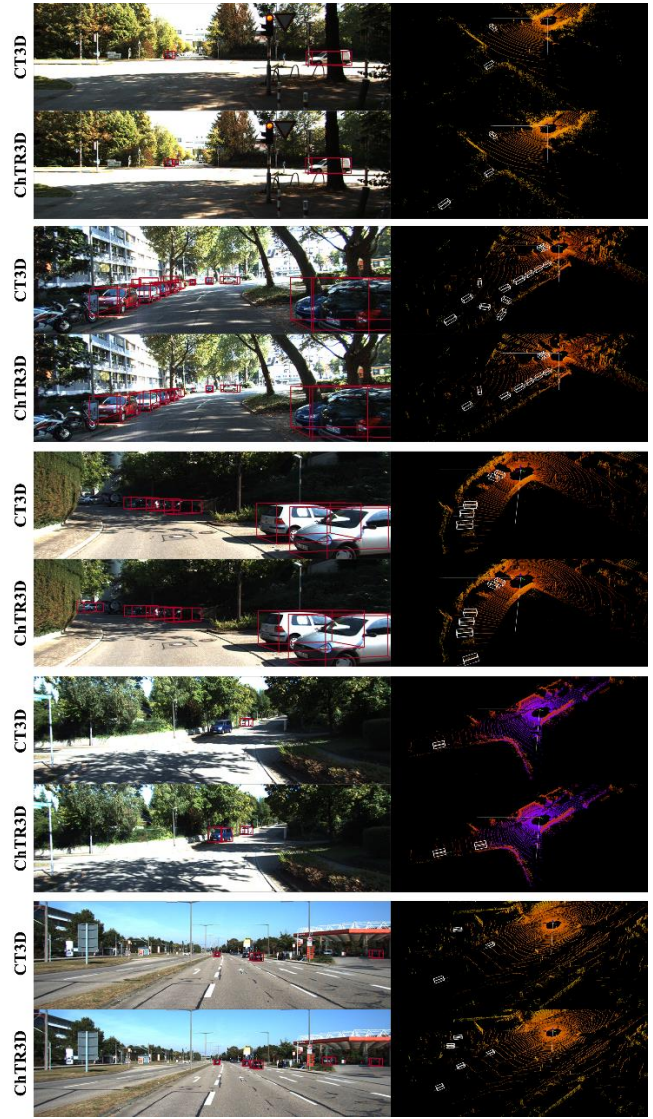

Figure 5. Visualization of CT3D and ChTR3D performance on the KITTI test set, the detection results of our ChTR3D are more reasonable.

TABLE III. PERFORMANCE OF DIFFERENT VALUES OF $a$ ON KITTI VALIDATION SET

| Method | Easy | Mod. | Hard | mAP |
|---|---|---|---|---|
| **ChTR3D** | | | | |
| $a = 1.3169$ | 91.87 | 84.88 | 82.73 | 86.49 |
| $a = 1.1$ | 92.01 | 84.98 | 82.86 | 86.62 |
| $a = 0.9$ | 91.14 | 85.03 | 82.71 | 86.29 |
| $a = 0.7$ | 91.48 | 84.26 | 82.62 | 86.12 |
| $a = 0.5$ | 92.38 | 85.36 | 83.12 | 86.95 |
| $a = 0.3$ | 92.10 | 84.87 | 82.69 | 86.55 |
| $a = 0.1$ | 91.75 | 84.84 | 82.93 | 86.51 |

The evaluated metric AP is calculated by 0.7 IoU threshold and 40 recall positions for *car* category.